\documentclass[letterpaper]{article}
\usepackage[preprint]{aaai2027}
\usepackage[hyphens]{url}
\usepackage{graphicx}
\usepackage{natbib}
\usepackage{caption}
\usepackage{amsmath}
\usepackage{booktabs}

\title{SCA: Segment-Wise CoT Compression with Answer Alignment}
\author{Ye Tian, Hongyu Lin}
\affiliations{}

\begin{document}

\maketitle

\begin{abstract}
Chain-of-thought (CoT) reasoning improves problem solving, but long think
traces increase inference cost. Existing CoT compression methods
usually optimize completion-level length. For structured thinking models,
however, a completion contains both a \textit{think} segment and an
\textit{answer} segment, so completion-level compression can save
tokens by compressing not only the CoT but also the answer. We call this failure mode
\textit{answer drift}. We propose \textbf{Segment-wise CoT Compression with
Answer Alignment (SCA)}, an answer-preserving think-compression method. SCA
parses completions into functional segments, routes compression rewards only to
successful \textit{think} tokens, and protects \textit{answer} tokens through
length and distribution alignment to a frozen base model. Experiments show that,
across datasets from multiple domains, SCA achieves state-of-the-art-level
chain-of-thought compression while preserving the base model's performance and
answer alignment. Training data and code are included in the supplementary
code and data package.
\end{abstract}

\section{Introduction}

Long CoT reasoning can improve the accuracy of language models on difficult
tasks, but it also increases inference cost. This motivates CoT
compression: after a model has learned to reason, we would like it to reach the
same result with shorter reasoning traces. Recent methods show that reasoning
traces can often be shortened while retaining final-answer accuracy
\cite{liu2024can,xia-etal-2025-tokenskip,li2026making,ma-etal-2025-cot,
liang2026deepcompress}.

In structured thinking models, however, a generated completion is not a
homogeneous string. The \textit{think} segment mainly carries computation, while
the \textit{answer} segment carries the response that is evaluated and read.
Because both segments are generated by the same policy, completion-level
compression can change answer behavior rather than only reduce reasoning cost.

\begin{figure*}[t]
\centering
\includegraphics[width=0.92\textwidth]{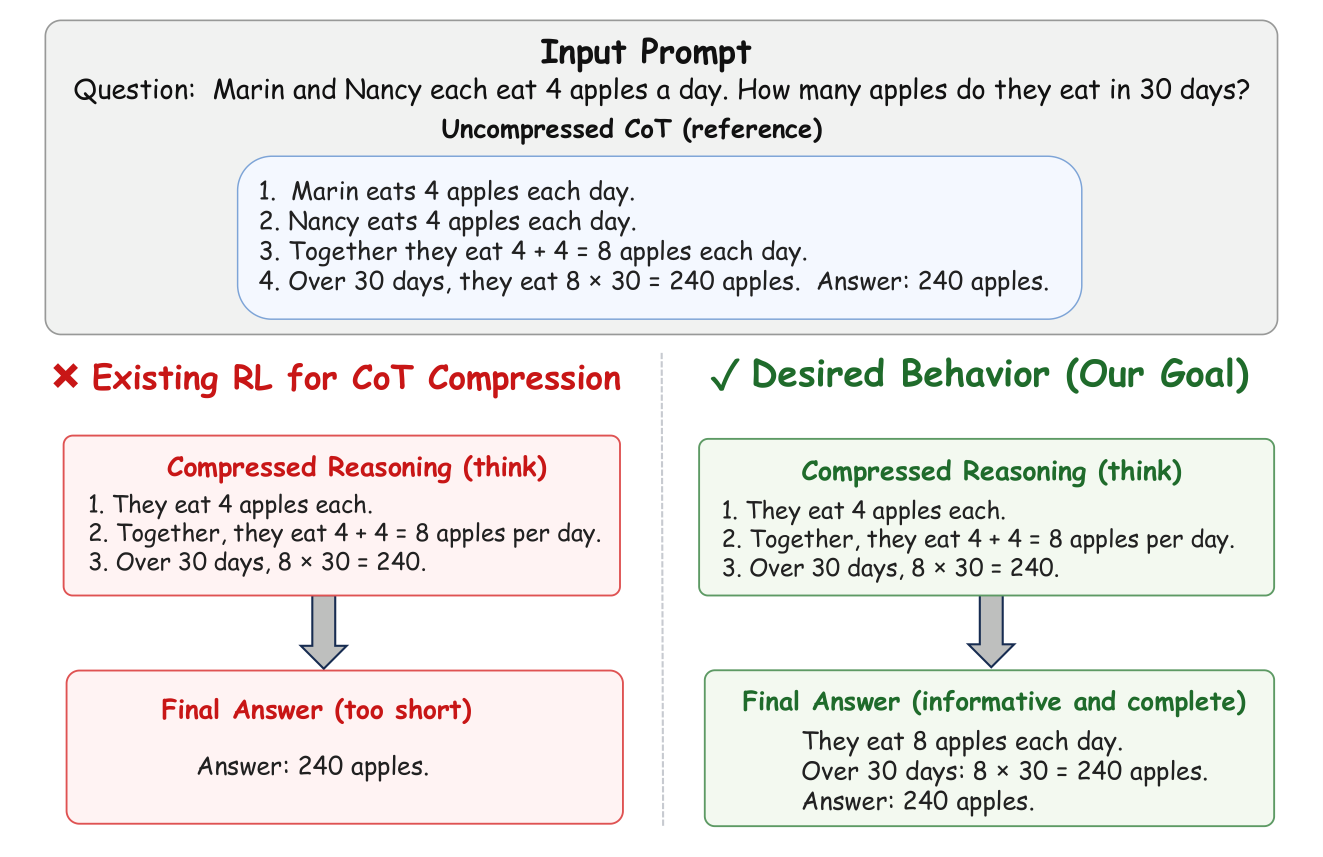}
\caption{Example of answer drift and the desired behavior of SCA.}
\label{fig:motivation}
\end{figure*}

Prior CoT compression work often evaluates compression mainly through accuracy
and total length, but these metrics do not fully describe what changed. Many
reasoning models emit structured outputs with a \textit{think} segment
and an \textit{answer} segment. If a training objective rewards shorter
whole completions, both reasoning tokens and answer tokens
can reduce the objective. The model may therefore keep the final result correct
while removing derivations or explanations that make the answer useful.
We refer to this behavior as \textit{answer drift}. Figure
\ref{fig:motivation} illustrates the problem: completion-level compression can
leave the final result intact but remove much of the answer-side explanation.

This paper studies \textbf{answer-preserving think compression}: given an
already-trained strong structured thinking model, reduce redundant
\textit{think} tokens while
keeping the \textit{answer} behavior close to the original base model.
This framing deliberately separates capability learning from cost reduction.
SCA is not intended to make an incapable model learn reasoning from scratch.
Instead, it is a post-training compression stage that can be applied to any
thinking model with structured outputs and already useful reasoning and answer
behavior.

\paragraph{Our Contributions.}
(1) \textbf{\textit{Answer drift.}} We identify answer drift in structured CoT
compression: completion-level compression can shorten the
\textit{answer} segment, not only the intended \textit{think} segment.
(2) \textbf{\textit{SCA.}} We propose \textbf{Segment-wise CoT Compression with
Answer Alignment (SCA)}, which separates \textit{think} and \textit{answer}
tokens during training. SCA routes compression to \textit{think} tokens,
preserves the \textit{answer} segment with answer alignment, and gives more
weight to successful \textit{think} trajectories on hard prompts.
(3) \textbf{\textit{Empirical results.}} Experiments demonstrate that SCA
effectively shortens \textit{think} traces, keeps answer-side
behavior close to the base model, and maintains competitive reasoning accuracy.

\section{Method}
\label{sec:method}

\begin{figure*}[t]
\centering
\includegraphics[width=0.95\textwidth]{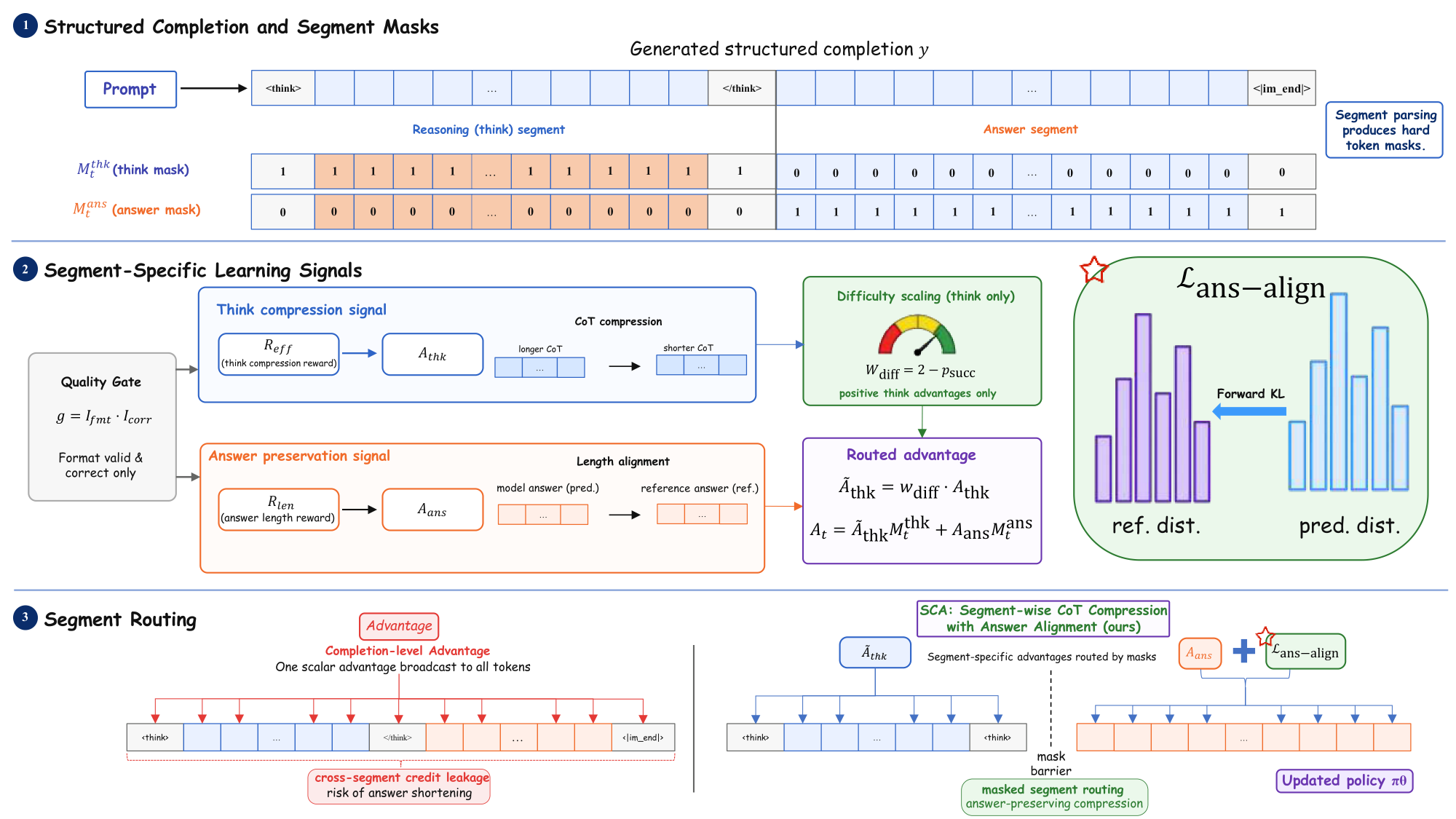}
\caption{Overview of SCA's segment-wise training pipeline.}
\label{fig:method_overview}
\end{figure*}

SCA parses each sampled completion into \textit{think} tokens and
\textit{answer} tokens, computes segment-specific rewards, and routes
the resulting token-level advantages back only to the corresponding segment.
Figure \ref{fig:method_overview} summarizes this pipeline. The key design
choice is that compression is not represented as a single completion-level
objective. Instead, the \textit{think} segment receives the efficiency signal,
while the \textit{answer} segment receives alignment signals from the frozen
base model.

\subsection{Segmentation and Quality Gate}

For each prompt $x$, we sample $K$ completions from the old policy. Each
completion is parsed using fixed boundary markers such as \texttt{<think>},
\texttt{</think>}, and an answer termination marker. The parser returns binary
masks $M_t^{\mathrm{thk}}(k)$ and $M_t^{\mathrm{ans}}(k)$ indicating whether
token $t$ in completion $k$ belongs to the \textit{think} segment or the
\textit{answer} segment. Segment lengths are
$L_{\mathrm{thk}}^{(k)}=\sum_t M_t^{\mathrm{thk}}(k)$ and
$L_{\mathrm{ans}}^{(k)}=\sum_t M_t^{\mathrm{ans}}(k)$.

Short malformed outputs should not be rewarded. We therefore use a quality gate
$g^{(k)}$, which equals 1 only when a completion is well formed and its answer
is judged equivalent to the gold answer. The format check requires
uniquely parseable think/answer boundaries and a non-empty answer. The gate is
applied to all structural rewards and to answer-side KL alignment.

\subsection{Think Compression Reward}

Among gated successful completions $\mathcal{G}=\{k:g^{(k)}=1\}$, SCA rewards
shorter reasoning relative to other successful rollouts for the same
prompt. Let $L_{\min}$ and $L_{\max}$ be the minimum and maximum successful
think lengths and let $\epsilon>0$ be a numerical stabilizer.
\begin{equation}
D=L_{\max}-L_{\min}+\epsilon,
\end{equation}
\begin{equation}
R_{\mathrm{eff}}^{(k)}
=
\begin{cases}
g^{(k)}, & |\mathcal{G}|\le 2,\\
g^{(k)}\phi(L_{\mathrm{thk}}^{(k)}), & |\mathcal{G}|>2,
\end{cases}
\end{equation}
\begin{equation}
\phi(L)=1-\dfrac{L-L_{\min}}{D}.
\end{equation}

\subsection{Answer Preservation}

The answer segment is protected with two complementary signals. First, SCA uses
a length-alignment reward. The reference answer length
$L_{\mathrm{ref}}(x)$ is the answer length produced by the frozen base
model for the same prompt. With tolerance $f$ and $U=L_{\mathrm{ref}}+f$,
\begin{align}
R_{\mathrm{len}}^{(k)}
&=g^{(k)}\rho(L_{\mathrm{ans}}^{(k)},L_{\mathrm{ref}}),\\
\rho(L,L_{\mathrm{ref}})
&=
\begin{cases}
\exp(-(L_{\mathrm{ref}}-L)/L_{\mathrm{ref}}), & L<L_{\mathrm{ref}},\\
1, & L_{\mathrm{ref}}\le L\le U,\\
\exp(-(L-U)/U), & L>U.
\end{cases}
\end{align}
This signal is a coarse guard against severe answer shortening. Second,
SCA applies forward KL on gated answer tokens:
\begin{align}
\mathcal{L}_{\mathrm{ans\text{-}align}}
&=
\frac{1}{Z_{\mathrm{ans}}}
\sum_{k,t}
g^{(k)}M_t^{\mathrm{ans}}(k)
\nonumber\\
&\quad
D_{\mathrm{KL}}\Big(
\pi_{\mathrm{ref}}(\cdot|s_t^{(k)})
\;\|\;
\pi_{\theta}(\cdot|s_t^{(k)})
\Big),
\end{align}
where $\pi_{\mathrm{ref}}$ is the same frozen base model and
$s_t^{(k)}=(x,y_{<t}^{(k)})$. We use forward KL because it treats the frozen
base model as the teacher distribution on answer tokens: the current policy is
penalized when it drops answer-token behavior that the base model assigns high
probability.
This term complements the length reward: length alignment constrains the
amount of answer text, while KL alignment preserves local token choices and
formatting patterns in the answer segment.

\begin{figure*}[!t]
\centering
\includegraphics[width=0.405\textwidth]{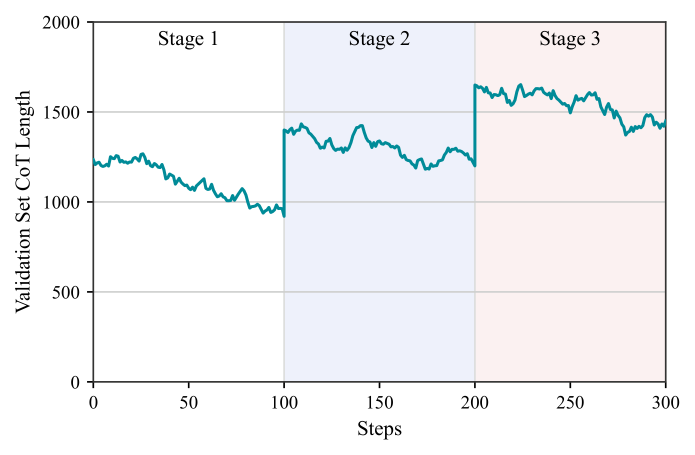}
\hfill
\includegraphics[width=0.405\textwidth]{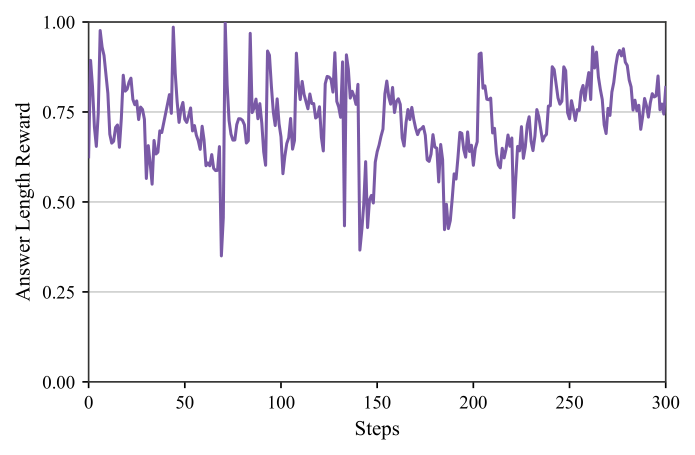}
\caption{Training dynamics for validation think length and answer-length
reward during SCA training.}
\label{fig:training_dynamics}
\end{figure*}

\begin{table*}[!t]
\centering
\small
\setlength{\tabcolsep}{4pt}
\renewcommand{\arraystretch}{0.80}
\begin{tabular*}{\textwidth}{@{\extracolsep{\fill}}llcccccc@{}}
\toprule
Model & Method & MATH-500 & AMC23 & MinervaMath & AIME24 & AIME25 & Avg. \\
\midrule
Qwen3-4B & Base & $97.2{\pm}0.3$ & $97.5{\pm}2.3$ & $69.9{\pm}0.4$ & $70.0{\pm}3.1$ & $73.3{\pm}3.1$ & 81.59 \\
Qwen3-4B & DeepCompress & $97.0{\pm}0.3$ & $97.5{\pm}2.3$ & $72.3{\pm}0.5$ & $70.0{\pm}3.1$ & $73.3{\pm}2.5$ & 82.02 \\
Qwen3-4B & TokenSkip & $97.0{\pm}0.3$ & $97.5{\pm}2.3$ & $71.7{\pm}0.5$ & $70.0{\pm}3.1$ & $73.3{\pm}3.1$ & 81.90 \\
Qwen3-4B & SCA & $97.2{\pm}0.1$ & $97.5{\pm}2.3$ & $71.1{\pm}0.3$ & $73.3{\pm}2.5$ & $73.3{\pm}3.1$ & 82.49 \\
\midrule
Qwen3-8B & Base & $97.0{\pm}0.3$ & $95.0{\pm}2.3$ & $72.1{\pm}0.5$ & $76.7{\pm}3.1$ & $66.7{\pm}2.5$ & 81.49 \\
Qwen3-8B & DeepCompress & $97.2{\pm}0.3$ & $97.5{\pm}1.9$ & $71.1{\pm}0.5$ & $73.3{\pm}3.1$ & $73.3{\pm}3.1$ & 82.49 \\
Qwen3-8B & TokenSkip & $97.2{\pm}0.3$ & $97.5{\pm}2.3$ & $71.7{\pm}0.5$ & $73.3{\pm}3.1$ & $73.3{\pm}2.5$ & 82.61 \\
Qwen3-8B & SCA & $97.4{\pm}0.3$ & $97.5{\pm}2.3$ & $72.3{\pm}0.5$ & $73.3{\pm}2.5$ & $73.3{\pm}3.1$ & 82.77 \\
\bottomrule
\end{tabular*}
\caption{Pass@1 accuracy under multiple decoding seeds on mathematical
benchmarks.}
\label{tab:main_acc}
\end{table*}

\subsection{Segment-Wise Objective}

SCA normalizes the think reward $R_{\mathrm{think}}=R_{\mathrm{eff}}$ and answer
reward $R_{\mathrm{answer}}=R_{\mathrm{len}}$ separately within the sampled
group. For $q=\mathrm{thk}$, $R_q=R_{\mathrm{think}}$ and
$\sigma_q=\sigma_{\mathrm{thk}}$; for $q=\mathrm{ans}$,
$R_q=R_{\mathrm{answer}}$ and $\sigma_q=\sigma_{\mathrm{ans}}$:
\begin{equation}
\mathrm{adv}_{q}^{(k)}
=
\frac{R_q^{(k)}-K^{-1}\sum_j R_q^{(j)}}{\sigma_q(x)+\epsilon},
\quad q\in\{\mathrm{thk},\mathrm{ans}\},
\end{equation}
where $\sigma_{\mathrm{thk}}(x)$ and $\sigma_{\mathrm{ans}}(x)$ are the
within-group standard deviations for the corresponding segment returns. Let
$\hat{p}_{\mathrm{succ}}(x)=K^{-1}\sum_k g^{(k)}$ be the observed prompt-level
success rate. Inspired by analyses of GRPO-style positive advantages
\cite{yao2026grouprelative,Oliveira2025LearningWC}, we define
\begin{equation}
W_{\mathrm{diff}}(x)=1+\alpha\bigl(1-\hat{p}_{\mathrm{succ}}(x)\bigr)
\end{equation}
where $\hat{p}_{\mathrm{succ}}(x)$ is an empirical difficulty estimate: harder
prompts have fewer successful rollouts and thus receive larger positive-think
credit. The linear form is monotone and bounded,
$W_{\mathrm{diff}}(x)\in[1,1+\alpha]$, so even very hard prompts cannot produce
unstable amplification. We then amplify only positive think advantages:
\begin{equation}
\widetilde{\mathrm{adv}}_{\mathrm{thk}}^{(k)}
=
\begin{cases}
\mathrm{adv}_{\mathrm{thk}}^{(k)}W_{\mathrm{diff}}(x),
& \mathrm{adv}_{\mathrm{thk}}^{(k)}>0,\\
\mathrm{adv}_{\mathrm{thk}}^{(k)}, & \text{otherwise}.
\end{cases}
\end{equation}
This addresses signal dilution on hard prompts: when successful gated rollouts
are rare, their useful think trajectories can otherwise be overwhelmed by many
heterogeneous failures. The amplification is asymmetric, so it strengthens only
success-aligned think compression without magnifying answer-side preservation
signals or noisy negative advantages from failures. This helps the model learn
more concise reasoning traces without paying for compression with a drop in
task performance.

The token-level advantage is routed by segment:
\begin{equation}
A_t^{(k)}
=
\widetilde{\mathrm{adv}}_{\mathrm{thk}}^{(k)}M_t^{\mathrm{thk}}(k)
+
\mathrm{adv}_{\mathrm{ans}}^{(k)}M_t^{\mathrm{ans}}(k).
\end{equation}
Following the DAPO clipped objective, let $r_t^{(k)}$ be the old-policy
likelihood ratio and let $\bar{r}_t^{(k)}$ be its clipped version under
decoupled lower and upper bounds $\epsilon_{\mathrm{low}}$ and
$\epsilon_{\mathrm{high}}$. SCA uses the routed advantage in
\begin{align}
\mathcal{L}_{\mathrm{clip}}(\theta)
&=
-\frac{1}{Z}
\sum_{k=1}^{K}
\sum_{t=1}^{T}
\min\Big(
r_t^{(k)}(\theta)A_t^{(k)},
\nonumber\\
&\hspace{3.5em}
\bar{r}_t^{(k)}(\theta)A_t^{(k)}
\Big),
\nonumber\\
\mathcal{L}_{\mathrm{SCA}}(\theta)
&=
\mathcal{L}_{\mathrm{clip}}(\theta)
+
\lambda_{\mathrm{align}}
\mathcal{L}_{\mathrm{ans\text{-}align}}(\theta).
\end{align}
Here $Z$ normalizes over the routed tokens. The final objective therefore
combines segment-wise compression with answer alignment.
Unlike completion-level compression, SCA does not broadcast one advantage to
all tokens; it uses hard routing for segment-wise compression and answer-side
alignment for preservation. Additional pseudocode and implementation notes are
provided in the supplementary material. The training-code archive is released
with the paper.

\begin{table*}[!t]
\centering
\small
\setlength{\tabcolsep}{4pt}
\renewcommand{\arraystretch}{0.80}
\begin{tabular*}{\textwidth}{@{\extracolsep{\fill}}llccccc@{}}
\toprule
Model & Method & MATH-500 & AMC23 & MinervaMath & AIME24 & AIME25 \\
\midrule
\multicolumn{7}{c}{Average think length} \\
\midrule
Qwen3-4B & Base & 3520 & 7333 & 3853 & 12215 & 15341 \\
Qwen3-4B & DeepCompress & 1861 & 4493 & 2555 & 8089 & 9758 \\
Qwen3-4B & TokenSkip & 2470 & 5150 & 2775 & 8750 & 11050 \\
Qwen3-4B & SCA & 1975 & 4831 & 2543 & 8192 & 9986 \\
\midrule
Qwen3-8B & Base & 3211 & 6543 & 3309 & 9854 & 14064 \\
Qwen3-8B & DeepCompress & 1668 & 3975 & 1936 & 7073 & 8869 \\
Qwen3-8B & TokenSkip & 2250 & 4580 & 2310 & 7600 & 9950 \\
Qwen3-8B & SCA & 1847 & 3729 & 1947 & 7174 & 8978 \\
\midrule
\multicolumn{7}{c}{Average answer length} \\
\midrule
Qwen3-4B & Base & 635 & 709 & 449 & 5099 & 5148 \\
Qwen3-4B & DeepCompress & 354 & 342 & 233 & 1451 & 1677 \\
Qwen3-4B & TokenSkip & 590 & 690 & 355 & 4650 & 4061 \\
Qwen3-4B & SCA & 620 & 754 & 531 & 4982 & 4554 \\
\midrule
Qwen3-8B & Base & 679 & 779 & 557 & 3208 & 1125 \\
Qwen3-8B & DeepCompress & 258 & 347 & 255 & 1492 & 510 \\
Qwen3-8B & TokenSkip & 581 & 634 & 511 & 3003 & 909 \\
Qwen3-8B & SCA & 614 & 738 & 579 & 3239 & 1006 \\
\bottomrule
\end{tabular*}
\caption{Average think and answer lengths on mathematical benchmarks.}
\label{tab:main_len}
\end{table*}

\section{Experiments}
\label{sec:experiments}

\begin{table*}[!t]
\centering
\small
\setlength{\tabcolsep}{6pt}
\begin{tabular*}{\textwidth}{@{\extracolsep{\fill}}llcccccc@{}}
\toprule
Model & Method & \multicolumn{2}{c}{Accuracy} & \multicolumn{2}{c}{Think Len.} & \multicolumn{2}{c}{Answer Len.} \\
\cmidrule(lr){3-4}\cmidrule(lr){5-6}\cmidrule(l){7-8}
 & & HumanEval & GPQA & HumanEval & GPQA & HumanEval & GPQA \\
\midrule
Qwen3-4B & Base & $87.80{\pm}0.80$ & $58.08{\pm}0.66$ & 4017 & 7568 & 162 & 943 \\
Qwen3-4B & SCA & $87.88{\pm}1.00$ & $58.65{\pm}0.83$ & 3095 & 4921 & 153 & 895 \\
Qwen3-8B & Base & $88.41{\pm}0.80$ & $59.09{\pm}0.76$ & 2360 & 7814 & 176 & 874 \\
Qwen3-8B & SCA & $88.49{\pm}1.00$ & $59.09{\pm}0.66$ & 1702 & 2916 & 166 & 832 \\
\bottomrule
\end{tabular*}
\caption{Cross-domain accuracy, think length, and answer length on HumanEval
and GPQA-Diamond.}
\label{tab:cross_domain}
\end{table*}

\subsection{Setup}

We evaluate Qwen3-4B-Thinking-2507 and Qwen3-8B
\cite{yang2025qwen3technicalreport}. These models are a natural testbed for
SCA because they already have strong task behavior, produce long think
traces, and use stable think/answer boundaries. SCA is therefore
used as an additional compression stage applied after capability learning, not
as a method for teaching a weak model to reason from scratch.

SCA is compared with the base model, a reproduced DeepCompress baseline
\cite{liang2026deepcompress}, and TokenSkip
\cite{xia-etal-2025-tokenskip}. DeepCompress and SCA use the same base models,
curriculum, rollout budget, and evaluation protocol; they differ mainly in
whether compression is optimized at completion level or routed to functional
segments. TokenSkip is added as a non-RL explicit CoT compression baseline that
can be evaluated under the same decoding protocol. This set of baselines does
not cover every latent, distillation, or architecture-changing compressor, but
it directly tests whether token savings come from reasoning or answer segments
under matched decoding.

Training uses Qwen3-4B-Thinking-2507 and Qwen3-8B with a three-stage curriculum
over GSM8K \cite{cobbe2021gsm8k} and DeepMath-103K \cite{he2026deepmathk}:
2,000 GSM8K examples; 1,400 GSM8K plus 600 DeepMath examples with difficulty
$\leq 4$; and 1,000 GSM8K plus 500 low- and 500 high-difficulty DeepMath
examples. The curriculum first exposes the model to simpler arithmetic where
short correct trajectories are easier to observe, then gradually introduces
harder mathematical reasoning while retaining easier examples. We train on 32
NVIDIA A800 GPUs with 80GB memory for 300 full-parameter steps, one 100-step
round per stage, with batch size 16, $K=32$ rollouts, AdamW, learning rate
$10^{-5}$, and rollout temperature annealed from 1.3 to 0.7. Unless stated
otherwise, SCA uses $f=32$ and $\alpha=0.5$.

Figure \ref{fig:training_dynamics} reports the training diagnostics for this
curriculum. Validation think length can jump at stage boundaries because the
validation split becomes harder, but the within-stage trend still decreases.
The answer-length reward is noisier, since it is computed from sampled
answers, yet it stays in a stable range while think traces shrink. This
supports the intended behavior: compression pressure primarily affects
reasoning rather than causing systematic answer shortening.

Main mathematical evaluation uses MATH-500 \cite{lightman2023lets}, AMC 2023
\cite{maa_amc12}, MinervaMath \cite{lewkowycz2022minerva}, and AIME 2024/2025
\cite{maa_aime}; cross-domain evaluation uses HumanEval and GPQA-Diamond. We
decode $N=8$ samples per problem at temperature 0.7 with top-$p=0.95$ and a
maximum of 32768 new tokens, then report Pass@1, think length, answer length,
answer quality, ablations, and diagnostics. Additional evaluation details,
judging prompts, and preprocessing rules are provided in the supplementary
material.

\subsection{Accuracy Retention and Segment-Level Compression}

Table \ref{tab:main_acc} reports Pass@1 with decoding-seed standard deviations.
Across both model scales, continuing training with DeepCompress, TokenSkip, or
SCA does not produce a clear degradation relative to the base model. The small
differences among methods are generally within decoding variation, especially
on the small AIME sets, so we treat these results as evidence of accuracy
retention rather than as significant improvement claims.

Table \ref{tab:main_len} shows where token reduction occurs. The key
difference among the three compression methods is not whether they can reduce
length, but which segment absorbs the reduction. DeepCompress and TokenSkip can
shorten generations, but their compression is less cleanly separated from the
answer segment. SCA is the only method in this comparison that both
substantially shortens thinking and keeps the answer close to
the base model's response length.
This distinction is the empirical signature of answer drift. If only total
length were measured, a method could look effective while obtaining part of its
token savings by deleting answer explanations. The segment-level view separates
these two effects: reducing redundant reasoning is desirable, whereas reducing
the answer segment changes the behavior that downstream users and evaluators
actually inspect.

\begin{figure*}[!t]
\centering
\includegraphics[width=0.96\textwidth]{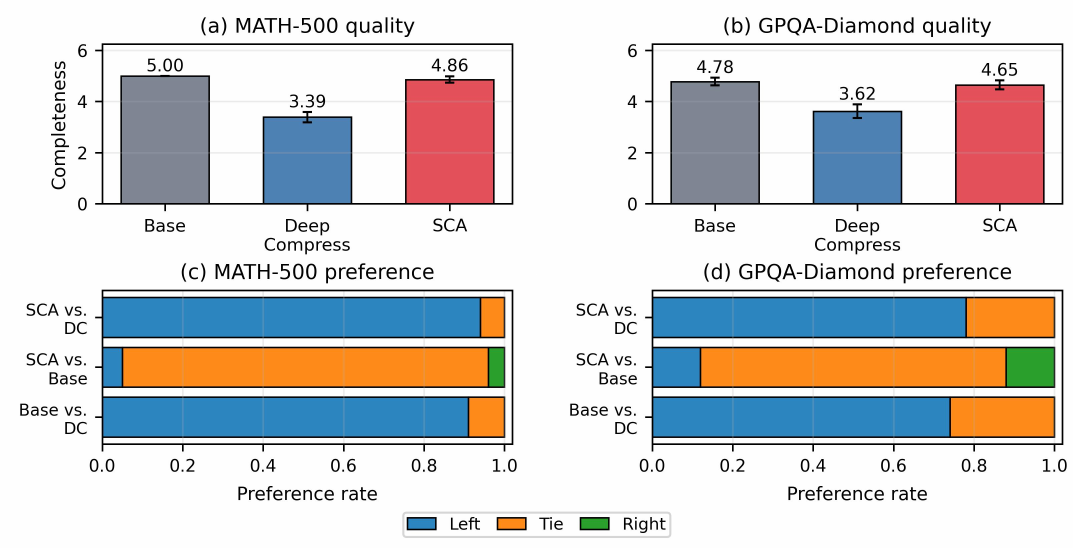}
\caption{Answer quality on final-answer-correct samples judged from the answer
segment only. DC denotes DeepCompress.}
\label{fig:answer_quality}
\end{figure*}

\subsection{Answer Quality and Cross-Domain Behavior}

Answer length is only a proxy for answer preservation. We therefore judge
answer segments on final-answer-correct samples, excluding think tokens from
the judge input. Figure \ref{fig:answer_quality} shows that DeepCompress can
keep the final answer while compressing visible explanations, whereas SCA
remains closer to the base model's answer style.

Table \ref{tab:cross_domain} shows the same segment-level pattern beyond
mathematics. On HumanEval \cite{chen2021evaluating} and GPQA-Diamond
\cite{rein2024gpqa}, SCA preserves performance, shortens think traces, and
keeps answer length close to the base model. Since these tasks differ from math
benchmarks in format, evidence structure, and required skills, the transfer
suggests that SCA learns a general control over reasoning allocation rather
than a math-specific compression shortcut.

\subsection{Ablations and Diagnostics}

Table \ref{tab:ablation} isolates the main components on GSM8K-test. Removing
difficulty weighting hurts accuracy, indicating that successful trajectories on
hard prompts need stronger positive credit. Removing all answer-side alignment
acts like a think-only compression baseline: the think reward remains routed,
but answer tokens receive no explicit protection, so the answer becomes much
shorter. The answer-side ablations show complementary roles: length alignment
guards against severe shortening, while KL preserves local answer-token
behavior; removing both yields the strongest answer degradation.

Table \ref{tab:hyperparam_diagnostics} reports a local sensitivity analysis
around the default SCA setting. The results show that SCA is not highly
sensitive to moderate changes in the answer tolerance or difficulty coefficient, and
the observed trends are consistent with the intended roles of these
hyperparameters. Increasing the answer tolerance from $f=32$ to $f=64$ relaxes
the no-penalty region around the base-model answer length, allowing the model
to retain slightly longer answers, but it does not improve accuracy. We
therefore use $f=32$ as the default, since it already keeps answer length close
to the base model while avoiding an unnecessarily loose preservation
constraint. Increasing the difficulty coefficient from $\alpha=0.5$ to
$\alpha=1.0$ gives
stronger credit to successful think trajectories on hard prompts. This slightly
improves accuracy but also increases think length, indicating a trade-off
between preserving more reasoning on difficult examples and achieving stronger
compression. We choose $\alpha=0.5$ as the default because it maintains
base-level accuracy while producing shorter think traces.

\begin{table}[!t]
\centering
\small
\setlength{\tabcolsep}{0pt}
\begin{tabular*}{\columnwidth}{@{\extracolsep{\fill}}lccc@{}}
\toprule
Method & Acc. & Think & Answer \\
\midrule
Base & 93.8 & 1467 & 130 \\
SCA & 93.8 & 617 & 133 \\
w/o Difficulty Weight & 89.8 & 603 & 130 \\
w/o Answer Alignment & 93.8 & 579 & 79 \\
w/o Segment Routing & 93.3 & 1447 & 132 \\
w/o Answer Length Reward & 93.2 & 615 & 97 \\
w/o Answer KL & 94.0 & 618 & 114 \\
\bottomrule
\end{tabular*}
\caption{Fine-grained component ablation on GSM8K-test.}
\label{tab:ablation}
\end{table}

\begin{table}[!t]
\centering
\small
\setlength{\tabcolsep}{0pt}
\begin{tabular*}{\columnwidth}{@{\extracolsep{\fill}}lccc@{}}
\toprule
Setting & Acc. & Think & Answer \\
\midrule
Default: $f=32$, $\alpha=0.5$ & 93.8 & 617 & 133 \\
$f=64$ & 93.4 & 631 & 138 \\
$\alpha=1.0$ & 94.3 & 654 & 129 \\
\bottomrule
\end{tabular*}
\caption{Hyperparameter diagnostics on GSM8K-test.}
\label{tab:hyperparam_diagnostics}
\end{table}

\subsection{Lightweight LoRA Diagnostic}

\begin{figure}[!t]
\centering
\includegraphics[width=0.88\columnwidth]{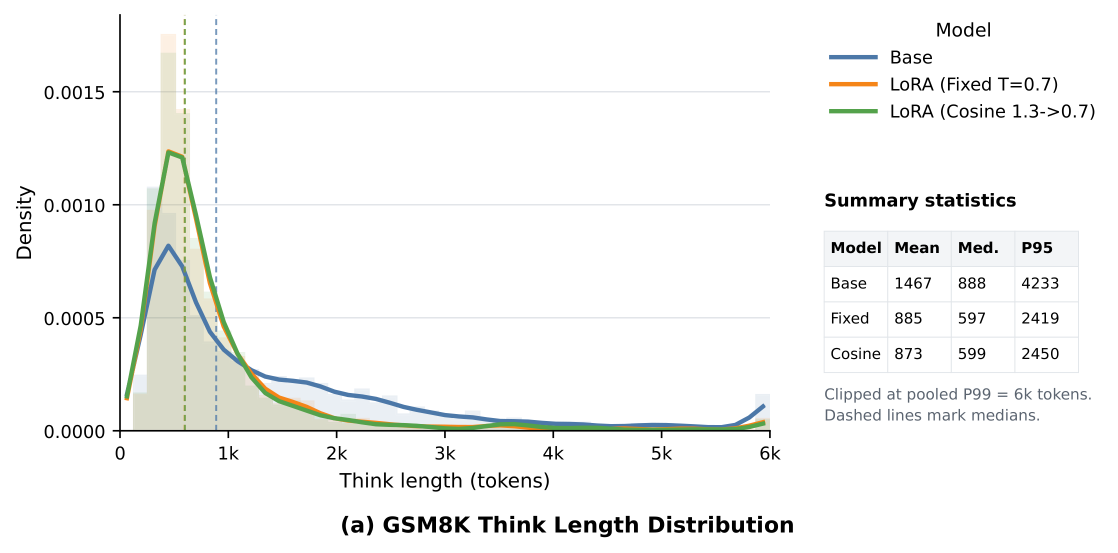}
\includegraphics[width=0.88\columnwidth]{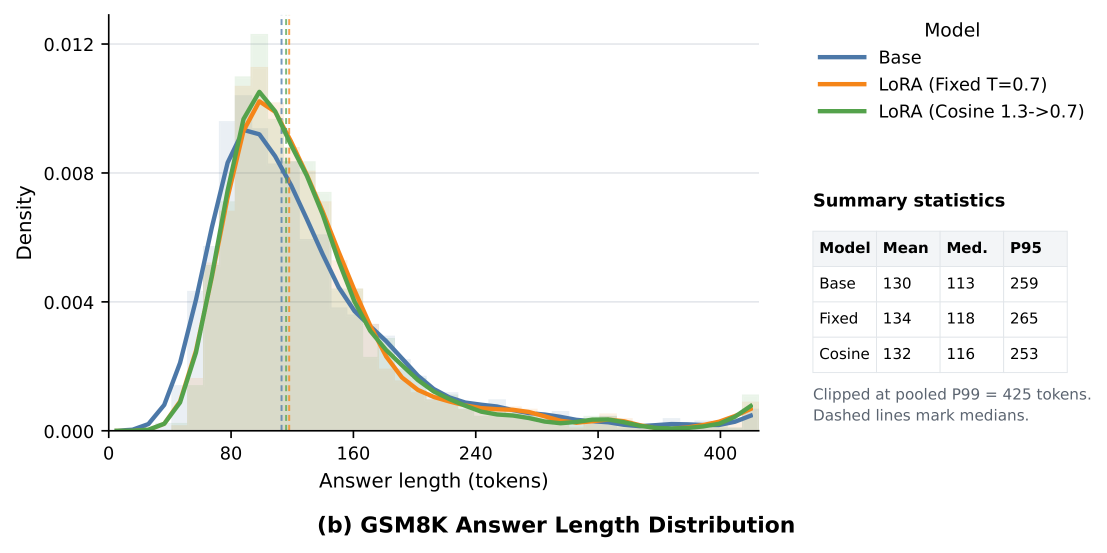}
\caption{GSM8K-test think-length and answer-length distributions for the
lightweight LoRA diagnostic at evaluation temperature 0.7.}
\label{fig:gsm8k_len_dists}
\end{figure}

We also include a lightweight GSM8K LoRA diagnostic to test whether small
adapter updates transfer beyond the in-domain GSM8K-test setting. Table
\ref{tab:lora_diagnostics} shows that LoRA produces only limited cross-benchmark
think-length reductions compared with full-parameter SCA. Figure
\ref{fig:gsm8k_len_dists} shows the corresponding in-domain distributions:
LoRA shifts GSM8K-test think lengths left, but it does not reliably remove the
long tail, and the transfer to harder mathematical benchmarks remains weak.
This suggests that answer-preserving reasoning compression is not merely a
surface preference for shorter text. The allocation of reasoning tokens reflects
deeper generation behavior distributed across many parameters, so it is not
reliably reshaped by small-adapter updates alone.

\begin{table}[!t]
\centering
\small
\setlength{\tabcolsep}{0pt}
\begin{tabular*}{\columnwidth}{@{\extracolsep{\fill}}lccccc@{}}
\toprule
Setting & MATH & AMC & Min. & A24 & A25 \\
\midrule
Base & 3520 & 7333 & 3853 & 12215 & 15341 \\
LoRA fixed & 3440 & 7160 & 3788 & 11980 & 15020 \\
LoRA cosine & 3375 & 7055 & 3710 & 11840 & 14860 \\
\bottomrule
\end{tabular*}
\caption{LoRA transfer diagnostics across mathematical benchmarks.}
\label{tab:lora_diagnostics}
\end{table}

\section{Related Work}

\textbf{Reasoning RL and long CoT.}
CoT improves LLM reasoning by exposing intermediate steps
\cite{chu2024navigate,li2024chain,ton2025understanding}. Reasoning-oriented
RL systems such as DAPO show that policy optimization can elicit complex
reasoning behavior at scale \cite{yu2026dapo}, while analyses of GRPO-style
objectives study how group-relative advantages support critic-free updates
\cite{yao2026grouprelative}. However, longer or more explicit reasoning is not
uniformly beneficial: CoT can reduce performance on tasks where extra thinking
misleads the model \cite{liu2025mind}. This motivates controlling reasoning
length rather than treating longer CoT as uniformly better. SCA follows this view:
it is applied after a model has acquired useful reasoning and answer behavior,
and it treats token allocation as a post-training behavior to be reshaped.
Unlike work that primarily tries to elicit more reasoning, our goal is to
retain the useful behavior of a strong thinking model while reducing redundant
reasoning after the fact. This makes the optimization target different from
capability learning: the training signal should not simply reward shorter
outputs, but should preserve the parts of the response that carry the answer.

\textbf{CoT compression and token-efficient reasoning.}
Token-efficient reasoning has been studied through latent reasoning
\cite{shen-etal-2025-codi,tan2026think,wei2026simcot}, explicit token or step
compression \cite{liu2024can,xia-etal-2025-tokenskip,li2026making,
ma-etal-2025-cot}, and distillation-based compact reasoning
\cite{chen2025skip,zhuang2025unicott,feng2026cotevo}. DeepCompress further
uses RL rewards to balance exploration, answer correctness, and reasoning
compression \cite{liang2026deepcompress}. These methods demonstrate that CoT
traces are compressible, but most report task performance together with
reasoning or completion length. For structured outputs, such metrics do not
reveal whether the answer segment is preserved. SCA therefore treats compression as a
segment-wise problem: the desired savings should come from redundant reasoning,
not from answer content that makes the final response checkable.
This perspective is complementary to prior compression mechanisms. Latent or
implicit reasoning changes where reasoning is represented, explicit skipping or
step compression changes how much text is generated, and distillation transfers
compact reasoning patterns from one model or trace distribution to another. SCA
instead keeps the structured output format and asks which segment should absorb
the compression pressure. This is especially important for models that expose a
separate answer segment, because a shorter completion is not necessarily a
better compressed reasoning process.

\textbf{Credit assignment for structured outputs.}
Group-relative RL removes learned critics by estimating advantages from
within-group comparisons, but recent analyses also show that group-relative
updates can introduce nontrivial optimization effects, including sensitivity to
baselines, grouping, and negative gradients
\cite{yao2026grouprelative,deng2026on,Oliveira2025LearningWC}. Step-level
methods address a related issue by assigning credit within reasoning traces
\cite{li2026ssvpo}. Our setting differs: the target is not only reasoning
steps, but functional segments with different objectives. SCA routes
compression to \textit{think} tokens and preservation to \textit{answer}
tokens, reducing cross-boundary credit leakage. This use of segment boundaries
is a lightweight alternative to dense token-level critics: the output format
itself defines where each training signal should act.
Compared with step-level credit assignment, the segment boundary is coarser but
directly tied to the objective mismatch in structured CoT compression. The
reasoning segment should become more efficient, while the answer segment should
remain aligned with the base model's response behavior. Treating these two
regions as the same optimization target can make token savings appear to come
from reasoning even when they partly come from answer shortening.

\section{Limitations and Conclusion}

\paragraph{Limitations.}
SCA is intended for structured thinking models that use stable think/answer
boundaries and for checkpoints that already have useful task behavior and a
reasonable answer style. Our experiments cover Qwen3 thinking models
across math, code, and expert QA benchmarks. The same segment-aware view can be
extended to other model families, richer response formats, and tool-use traces,
as long as a reasoning trace can be separated from the answer segment.

\paragraph{Conclusion.}
We introduced answer drift: completion-level CoT compression can save tokens by
removing useful answer content. More generally, structured outputs should be
treated as structured objects during both optimization and evaluation, because
equal token savings can have different effects depending on whether they come
from the think segment or the answer segment.
SCA addresses this mismatch with a segment-wise training objective. It routes
compression pressure to successful think tokens and uses answer-side length and
distribution alignment to keep the final response close to the frozen base
model. Across mathematical, code, and expert QA benchmarks, SCA preserves task
performance while shortening think traces and keeping answer behavior closer to
the base model than completion-level compression.

\bibliography{references}

\end{document}